# A theoretical study of Y structures for causal discovery

**Subramani Mani**\*
mani@uwm.edu
Department of EECS
University of Wisconsin-Milwaukee
Milwaukee, WI 53211

**Peter Spirtes**
ps7z@andrew.cmu.edu
Department of Philosophy
Carnegie Mellon University
Pittsburgh, PA 15213

**Gregory F. Cooper**
gfc@cbmi.pitt.edu
Center for Biomedical Informatics
University of Pittsburgh
Pittsburgh, PA 15213

## Abstract

Causal discovery from observational data in the presence of unobserved variables is challenging. Identification of so-called Y substructures is a sufficient condition for ascertaining some causal relations in the large sample limit, without the assumption of no hidden common causes. An example of a Y substructure is $A \to C, B \to C, C \to D$. This paper describes the first asymptotically reliable and computationally feasible score-based search for discrete Y structures that does not assume that there are no unobserved common causes. For any parameterization of a directed acyclic graph (DAG) that has scores with the property that any DAG that can represent the distribution beats any DAG that can't, and for two DAGs that represent the distribution, if one has fewer parameters than the other, the one with the fewest parameter wins. In this framework there is no need to assign scores to causal structures with unobserved common causes. The paper also describes how the existence of a Y structure shows the presence of an unconfounded causal relation, without assuming that there are no hidden common causes.

## 1 Introduction

Discovering causal relationships from observational data is challenging due to the presence of observed confounders[1], particularly, hidden (latent) confounders.

---

\*Currently at the Department of Biomedical Informatics, Vanderbilt University, Nashville, TN 37232-8340.

[1] A node $W$ is said to be a confounder of nodes $A$ and $B$ if there is a directed path from $W$ to $A$ and a directed path from $W$ to $B$ that does not traverse $A$. If $W$ is observed, it is said to be a measured confounder, otherwise it is a hidden confounder.

Furthermore, it is known that members of an independence (Markov) equivalence class of causal Bayesian network (CBN) models are indistinguishable using only probabilistic dependence and independence relationships among the observed variables.

There are several algorithms for reliably identifying (some) causal effects in the large sample limit, given the assumptions of acyclicity, and the Causal Markov and Causal Faithfulness assumptions (explained below). Due to space limitations, we mention only two representative algorithms here. The Greedy Equivalence Search (Chickering, 2002) is a score-based search that reliably identifies some causal effects in the large sample limit, but only under the additional assumption of no unobserved common causes. The FCI algorithm (Spirtes et al., 1999) reliably identifies some causal effects in the large sample limit, but it is a constraint-based search, and such methods have several important disadvantages relative to score-based searches (Heckerman et al., 1999)—(1) the inability to include prior belief in the form of structure probabilities, (2) the output is based on the significance level used for the independence tests, and (3) there is no quantification of the strength of the hypothesis that is output. A constraint based search performs a sequence of conditional independence tests. A single conditional independence test may force the removal of an edge, or a particular orientation, even if that makes the rest of the model a poor fit. In contrast, a score gives a kind of overall evaluation of how all the conditional independence constraints are met by a directed acyclic graph (DAG), and would not decrease the overall fit to save one conditional independence.

A score based search over models that explicitly include hidden variables faces several difficult problems: there are a potentially infinite number of such models, and there are both theoretical and computational difficulties in calculating scores for models with hidden variables (Richardson & Spirtes, 2002). Another approach is to use graphical models (mixed ances-tral graphs) that represent the marginal distributions of hidden variable models over the observed variables, but do not explicitly contain hidden variables (Richardson & Spirtes, 2002). Mixed ancestral graph models, unlike hidden variable models, are finite in number, and in the linear case, there are no theoretical or computational problems in scoring mixed ancestral graphs. However, there is no currently known way to score discrete mixed ancestral graphs, and there is no known algorithm for efficiently searching the space of mixed ancestral graphs. Finally, hidden variable models are known to entail non-conditional independence constraints upon the marginal distribution; these constraints can be used to guide searches for hidden variable models that entail the constraints (Spirtes et al., 1993; Tian & Pearl, 2002). However, no such computationally feasible search is known to be guaranteed correct.

The search proposed here is the first computationally feasible score-based search to reliably identify some causal effects in the large sample limit for both discrete and linear models, without assuming that there are no unobserved common causes, and without making any assumptions about the true causal structure (other than acyclicity). There is also no need to assign scores explicitly to causal structures with unobserved common causes in this framework. We identify a class of structures (Y structures) as "sufficient" for assigning causality, and provide the necessary theorems and proofs for the claim. The proofs presented in this paper depend only upon two features of the score—(1) a DAG that cannot represent a distribution loses in the limit to one that can; and (2) if two DAGs with different numbers of parameters can represent the distribution, the one with more parameters loses to the DAG with fewer parameters.

The remainder of the paper is organized as follows. Section 2 introduces the causal Bayesian network framework and presents the basic assumptions related to causal discovery. Section 3 describes the V and Y structures. Section 4 introduces the relevant theorems and proofs. Additional information on search and the score function is provided in Appendix A and Appendix B.

## 2 Background

In this section we first introduce the CBN framework. Subsequently, we define d-separation and d-connectivity, independence equivalence, the causal sufficiency assumption, the causal Markov condition, and the causal Faithfulness condition. The following notational convention will be used throughout the paper. Sets of variables are represented in bold and upper case and random variables by upper case letters italicized. Graphs are denoted by upper case letters such as G or M or by calligraphic letters such as $\mathcal{G}$ or $\mathcal{M}$. Lower case letters such as $x$ or $z$ denote an assignment to variables $X$ or $Z$.

### 2.1 Causal Bayesian network

A causal Bayesian network (CBN) is a directed acyclic graph (DAG) with the vertices representing observed variables in a domain and each arc is interpreted as a direct causal influence between a parent node[2] (variable) and a child node (Pearl, 1991). For example, if there is a directed edge from $A$ to $B$ ($A \rightarrow B$), node $A$ is said to exert a causal influence on node $B$. Figure 1 illustrates the structure of a hypothetical causal Bayesian network structure, which contains five nodes. The states of the nodes and the probabilities that are associated with this structure are not shown.

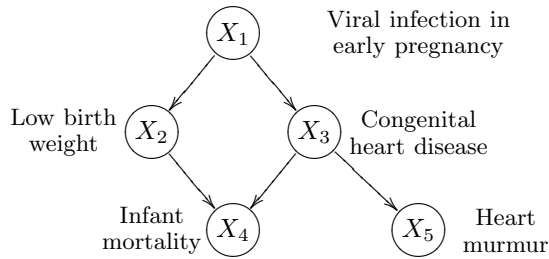

Figure 1: A hypothetical causal Bayesian network structure

The causal Bayesian network structure in Figure 1 indicates, for example, that a *Viral infection in early pregnancy* can causally influence whether *Congenital heart disease* is present in the newborn, which in turn can causally influence *Infant mortality* and presence of a *Heart murmur*.

### 2.2 d-separation and d-connectivity (Pearl, 1991)

d-separation is a graphical condition. Assume that $A$ and $B$ are vertices in a DAG $\mathcal{G}$ and $\mathbf{C}$ is a set of vertices in $\mathcal{G}$ such that $A \notin \mathbf{C}$ and $B \notin \mathbf{C}$. The d-separation theorem states that if a distribution P satisfies the Markov condition (each vertex is independent of its non-descendants conditional on its parents) for a DAG $\mathcal{G}$, and $A$ is d-separated from $B$ conditional on $\mathbf{C}$ in $\mathcal{G}$, then $A$ is independent of $B$ conditional on $\mathbf{C}$ in $\mathcal{G}$. Consider the DAG $\mathcal{G}$ in Figure 1. $A$ and $B$ are said to be d-separated given $\mathbf{C}$ iff the following property

---
[2] If there is an arc from node $A$ to node $B$ in a CBN, $A$ is said to be the parent of $B$, and $B$, the child of $A$.

holds: there exists no *undirected path*[3] $U$ between $X$ and $Y$ s.t.

1. every collider[4] on $U$ is in **C** or has a descendant in **C**.
2. no other vertex on $U$ is in **C**.

Likewise, if $A$ and $B$ are not in **C**, then $A$ and $B$ are *d-connected* given **C** iff they are not d-separated given **C**.

In Figure 1 the nodes $X_1$ and $X_5$ are d-separated by $X_3$. The nodes $X_2$ and $X_5$ are d-connected given {}. Two disjoint sets of variables **A** and **B** are d-separated conditional on **C** if and only if every vertex in **A** is d-separated from every vertex in **B** conditional on **C**; otherwise they are d-connected. See (Pearl, 1991) for more details on d-separation and d-connectivity.

### 2.3 Independence equivalence

Two Bayesian network structures S and S* over a set of observed variables **V** are independence equivalent (Heckerman, 1995) iff S and S* have the same set of d-separation and d-connectivity relationships between $A$ and $B$ conditional on **C** where $A \in \mathbf{V}$, $B \in \mathbf{V}$, and $\mathbf{C} \subset \mathbf{V}$ and $A \notin \mathbf{C}$ and $B \notin \mathbf{C}$. Independence equivalence is also referred to as Markov equivalence. For Gaussian or discrete distributions, it is also the case that two DAGs are independence equivalent if and only if the set of distributions that satisfy the Markov condition for one equals the set of distributions that satisfy the Markov condition for the other (i.e. they represent the same set of distributions).

### 2.4 The causal sufficiency assumption

A set of variables **S** is causally sufficient if no variable that is a direct cause[5] of two variables in **S** is not in **S**. In Figure 1, $\mathbf{S} = \{X_2, X_3\}$ is not causally sufficient because $X_1$ is a direct cause of $X_2$ and $X_3$, but is not in **S**. However, note that for the causal discovery approach based on Y structures introduced in this paper, we do not assume causal sufficiency.

### 2.5 The causal Markov condition

The **causal Markov condition** (CMC) gives the independence relationships that are specified by a causal Bayesian network:

> In a causal DAG for a causally sufficient set of variables, each variable is independent of its non-descendants (i.e., non-effects) given just its parents (i.e., its direct causes).

The CMC represents the "locality" of causality. This implies that indirect (distant) causes become irrelevant when the direct (near) causes are known. The CMC is the fundamental principle relating causal relations to probability distributions, and is explicitly or implicitly assumed by all causal graph search algorithms.

### 2.6 The causal Faithfulness condition

While the causal Markov condition specifies independence relationships among variables, the **causal Faithfulness condition** (CFC) specifies *dependence* relationships:

> In a causal DAG, two disjoint sets of variables **A** and **B** are dependent conditional on a third disjoint set of variables **C** unless the Causal Markov Condition entails that **A** and **B** are independent conditional on **C**.

The CFC is related to the notion that causal events are typically correlated in observational data. The CFC relates causal structure to probabilistic dependence. The CFC can fail for certain parameter values, but for linear or discrete parameters, the Lebesgue measure of the set of parameters for which it fails is 0. It is at least implicitly assumed by both constraint based and score based causal DAG discovery algorithms.

Based on the causal Markov condition each vertex of a CBN is independent of its non-descendants given its parents. The *independence map* or I-map of a CBN is the set of all independencies that follow from the causal Markov condition. The *dependence map* or D-map of a CBN is the set of all dependencies that follow from the causal Faithfulness condition.

The I-map and D-map of the causal network $W_1 \rightarrow X \leftarrow W_2$ is as follows:

- $W_1 \perp\!\!\!\perp W_2$; $W_1 \not\perp\!\!\!\perp X$; $W_1 \not\perp\!\!\!\perp X | W_2$
- $W_2 \not\perp\!\!\!\perp X$; $W_2 \not\perp\!\!\!\perp X | W_1$; $W_1 \not\perp\!\!\!\perp W_2 | X$

### 2.7 Markov Blanket

The Markov blanket (MB) of a node $X$ in a causal Bayesian network G is the union of the set of parents of $X$, the children of $X$, and the parents of the children of $X$ (Pearl, 1991). Note that the MB is the *minimal* set of nodes when conditioned on (instantiated) that makes a node $X$ independent of all the other nodes in the CBN.

---

[3]An undirected path between two vertices $A$ and $B$ in a graph $G$ is a sequence of vertices starting with $A$ and ending with $B$ and for every pair of vertices $X$ and $Y$ in the sequence that are adjacent there is an edge between them ($X \rightarrow Y$ or $X \leftarrow Y$) (Spirtes et al., 2000, page 8)

[4]A node with a head to head configuration. $C$ is a collider in $A \rightarrow C \leftarrow B$.

[5]The direct causation is relative to **S**.

## 3 V and Y structures

In this section we first define the V and Y structures and discuss their potential role in causal discovery.

### 3.1 V structure

A "V" structure over variables $W_1$, $W_2$ and $X$ is shown in DAG $M_1$ in Figure 2. There is a directed edge from $W_1$ to $X$ and another directed edge from $W_2$ to $X$. There is no edge between $W_1$ and $W_2$. A "V" structure contains a "collider" and the node $X$ is a collider in Figure 2, $M_1$. Since there is no arc between $W_1$ and $W_2$, $X$ is also termed as an *unshielded* collider. Figure 2, $M_2$ is a model in which there is an arc between $W_1$ and $W_2$, and thus $X$ is a *shielded* collider in this example.

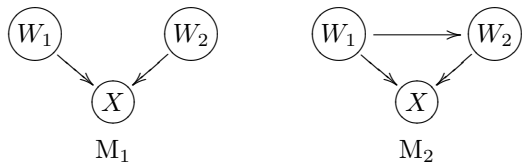

Figure 2: $X$ is an unshielded collider in $M_1$ and a shielded collider in $M_2$. $M_1$ is also referred to as a "V" structure.

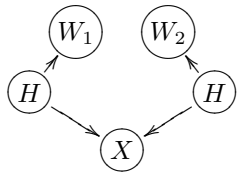

Figure 3: A model that has the same dependence/independence relationships over $W_1$, $W_2$ and $X$ as Figure 2, $M_1$; $H$ denotes a hidden variable.

The V structure is not *sufficient* for discovering that some variable $W_1$ or $W_2$ causes $X$ if we do not make the assumption that $M_1$ is causally sufficient. On the other hand, even allowing for confounders (hidden and measured), we can conclude that $X$ *does not* cause $W_1$ or $W_2$ (Spirtes et al., 2000). Figure 3 shows a confounder for the pair $(W_1, X)$ and the pair $(W_2, X)$. In other words we can make an *acausal* discovery but not a causal one using the V structure.

### 3.2 Y structure

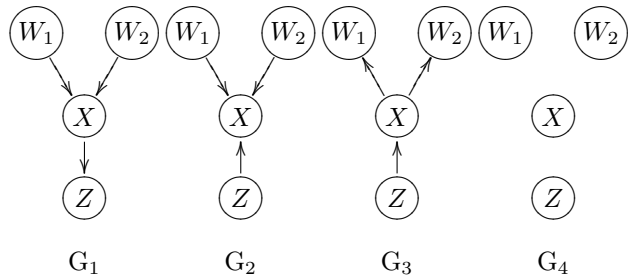

Figure 4: Several CBN models that contain four nodes. $G_1$ is a Y structure.

We now introduce the concept of a Y structure. Let $W_1 \to X \leftarrow W_2$ be a V structure. Note that $X$ is an unshielded collider in this V structure since there is no arc between $W_1$ and $W_2$. If there is a node $Z$ such that there is an arc from $X$ to $Z$, and there are no arcs from $W_1$ to $Z$ and $W_2$ to $Z$, then the nodes $W_1, W_2, X$ and $Z$ form a Y structure. A Y structure has interesting dependence and independence properties.

If $W_1, W_2, X, Z$ form a Y structure over a set of four variables $\mathbf{V}$ and the Y structure is represented by $G_1$ (see Figure 4), there is no DAG that contains a superset of the variables in $\mathbf{V}$, entails the same conditional independence relations over $\mathbf{V}$, and in which there is a (measured or unmeasured) confounder of $X$ and $Z$, or in which $X$ is not an ancestor of $Z$ (Robins et al., 2003; Spirtes et al., 2000, page 187) that is in the same independence equivalence class as $G_1$. In other words, if a Y structure is learned from data, the arc from $X$ to $Z$ represents an unconfounded causal relationship. Since $G_1$ also has the same set of independence/dependence relationships over the observed variables $W_1, W_2$, and $X$ as Figure 3), the arcs $W_1 \to X$ and $W_2 \to X$ cannot be interpreted as causal relationships.

## 4 Y structure theorems

**Definition 1 (Complete-table Bayesian network).** A complete-table Bayesian network is one that contains all discrete variables and for which the probabilities that define the Bayesian network are described by contingency tables with no missing values.

**Definition 2 (Perfect map).** A Bayesian network structure S is a perfect map of a distribution $\theta$ if $A$ and $B$ are independent conditional on $\mathbf{C}$ in $\theta$ iff $A$ and $B$ are d-separated conditional on $\mathbf{C}$ in S.

Suppose Bayesian network B defines a joint distribution $\theta$ over all the variables in B. Let S be the structure of B. If the Markov and Faithfulness conditions hold for B, then S is a perfect map of $\theta$.

In the results of this section, we will only be considering complete-table Bayesian networks that satisfy the Markov and Faithfulness conditions.

**Definition 3 (Y structure Bayesian network).** A Y structure Bayesian network is a Bayesian network containing four variables that has the structure shown in Figure 5, where the node labels are arbitrary.

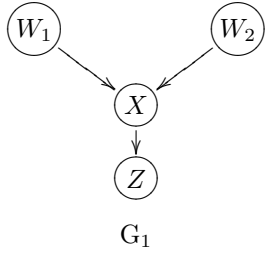

$G_1$

Figure 5: A Y structure.

We will use the following notation in regard to an arbitrary complete-table Bayesian network that satisfies the Markov and Faithfulness conditions and has a Y structure: $B_y$ denotes the network, $S_y$ denotes its structure, $\mathbf{V}_y$ denotes the four variables in the structure, $Q_y$ denotes its complete table parameters, and $\theta_y$ denotes the correspondingly defined joint distribution over the four variables.

**Lemma 1.** *There is no other Bayesian network structure on the variables in $\mathbf{V}_y$ that is independence equivalent to $S_y$.*

*Proof.* The proof of Lemma 1 follows from Theorem 1 given below.

**Theorem 1.** *Two network structures $B_{s1}$ and $B_{s2}$ are independence equivalent iff they satisfy the following conditions (Verma & Pearl, 1991):*

1. *$B_{s1}$ and $B_{s2}$ have the same set of vertices.*
2. *$B_{s1}$ and $B_{s2}$ have the same set of adjacencies.*
3. *If there is a configuration such as $A \to C \leftarrow B$ where $A$ and $B$ are not adjacent ("V" structure) in $B_{s1}$, the same pattern is present in $B_{s2}$, and vice-versa.*

□

**Lemma 2.** *Let B be a Bayesian network that contains the fewest number of parameters that can represent the population distribution. Let $B^*$ be a Bayesian network that either cannot represent the population distribution, or can but does not contain the fewest number of parameters. Let S and $S^*$ be the network structures of B and $B^*$, respectively. Let m denote the number of iid cases in D that have been sampled from the population distribution defined by B.*

$$\text{Then } \lim_{m \to \infty} \frac{P(S^*, D)}{P(S, D)} < 1. \quad (1)$$

*Proof.* The proof of Lemma 2 follows from the results in (Chickering, 2002), which in turn uses results in (Haughton, 1988). □

**Theorem 2.** *Let $B = (S, Q)$ be a complete-table Bayesian network that contains n measured variables, where S and Q are the structure and parameters of B, respectively. Suppose that B defines a distribution $\theta$ on the n variables, such that S is a perfect map of $\theta$. Let D be a database containing m complete cases on the n variables in B, for which the cases are iid samples from distribution $\theta$. Let $B^*$ be a Bayesian network over the same n variables with structure $S^*$ that is not independence equivalent to B. Suppose that $P(S, D)$ and $P(S^*, D)$ are computed using the BDe score with non-zero parameter and structure priors. (The BDe score assigns the same score to members of the same Markov equivalent class when equal structural priors are assigned; see (Heckerman et al., 1995).)*

$$\text{Then } \lim_{m \to \infty} \frac{P(S^*, D)}{P(S, D)} < 1. \quad (2)$$

*Proof.* If $B^*$ cannot represent the generative distribution $\theta$ then according to Lemma 2 the current theorem holds. Suppose $B^*$ can represent the generative distribution. Since by assumption $B^*$ is not independence equivalent to B, $B^*$ must contain all the dependence relationships in B, plus additional dependence relationships. Therefore $B^*$ contains more parameters than B (Chickering, 2002, Proposition 8). Thus it follows from Lemma 2 that the theorem holds. □

**Theorem 3.** *Assume the notation and conditions in Theorem 2 and suppose the number of variables is four (n = 4). If S is the data generating structure on the four variables and S is a Y structure, then in the large sample limit $P(S, D) > P(S^*, D)$ for all $S^* \neq S$ (where $S^*$ contains just the same 4 variables). Conversely, if S is the data generating structure and S is not equal to some Y structure, $S^*$, then in the large sample limit $P(S, D) > P(S^*, D)$.*

*Proof.* The proof follows from Theorem 2 and Lemma 1. □

Theorem 3 shows (under the conditions assumed) that in the large sample limit a Y structure will have the highest BDe score if and only if it is the structure of the data generating Bayesian network. Note that

Theorem 3 assumes that some DAG over the measured variables is a perfect map of the observed conditional independence and dependence relations; this is not in general true if there are unmeasured common causes.

Lemma 2 can be strengthened using results in (Nishii, 1988, Theorem 4) to show that in the large sample limit, with probability 1 the ratio approaches 0, rather than merely approaching some positive number less than 1.[6] Correspondingly, Theorem 3 can be strengthened to state that the data generating structure has posterior probability 1 and all other structures have probability 0 in the large sample limit[7]. This strengthened version of Theorem 3 implies that in the large sample limit, model averaging using Equation 3 in the appendix will derive an arc $X \rightarrow Z$ as causal and unconfounded with probability 1, if and only if it ($X \rightarrow Z$) is a causal and unconfounded arc within a Y structure of the data generating causal Bayesian network.

The proof of sufficiency of the Y structure for ascertaining causality in the possible presence of hidden confounders requires an understanding of the common properties of DAGs in the same Markov equivalence class in the possible presence of such hidden variables. A class of structures that can represent the common properties of DAGs in the same Markov equivalence class are called partial ancestral graphs (PAGs). We next describe the theorems that make use of PAGs. A PAG has a richer representation compared to a directed acyclic graph (DAG) and makes use of the following types of edges: $\rightarrow$, $\leftrightarrow$, $\circ\!\!\rightarrow$, $\circ\!\!-\!\!\circ$.

**Partial ancestral graphs**

A Markov equivalence class of DAGs over a set of observed variables $\mathbf{O}$ is the set of all DAGs that contain at least the variables in $\mathbf{O}$ and that have the same set of d-separation relations among the variables in $\mathbf{O}$ (i.e., $G_1$ and $G_2$ are in the same Markov equivalence class over $\mathbf{O}$ if for all disjoint $\mathbf{X}, \mathbf{Y}, \mathbf{Z} \subseteq \mathbf{O}$, $\mathbf{X}$ is d-separated from $\mathbf{Y}$ conditional on $\mathbf{Z}$ in $G_1$ iff $\mathbf{X}$ is d-separated from $\mathbf{Y}$ conditional on $\mathbf{Z}$ in $G_2$). A PAG $\mathcal{P}$ over $\mathbf{O}$ is a graphical object with vertices $\mathbf{O}$ that represents the Markov equivalence class of DAGs $\mathcal{M}$ over $\mathbf{O}$ in two distinct ways:

1. A PAG represents the d-separation relations over $\mathbf{O}$ in $\mathcal{M}$.

2. A PAG represents the ancestor and non-ancestor relations among variables in $\mathbf{O}$ common to every DAG in $\mathcal{M}$.

More specifically, it is possible to extend the concept of d-separation in a natural way to PAGs so that if PAG $\mathcal{P}$ represents the Markov equivalence class $\mathcal{M}$ over $\mathbf{O}$, then for all disjoint $\mathbf{X}, \mathbf{Y}, \mathbf{Z} \subseteq \mathbf{O}$, $\mathbf{X}$ is d-separated from $\mathbf{Y}$ conditional on $\mathbf{Z}$ in $\mathcal{P}$ iff $\mathbf{X}$ is d-separated from $\mathbf{Y}$ conditional on $\mathbf{Z}$ in every DAG in $\mathcal{M}$. A PAG is formally defined as stated below.

**Definition 4 (PAG).** The PAG $\mathcal{P}$ that represents a Markov equivalence class $\mathcal{M}$ over $\mathbf{O}$ can be formed in the following way:

1. $A$ and $B$ are adjacent in $\mathcal{P}$ iff $A$ and $B$ are d-connected conditional on every subset of $\mathbf{O}\backslash\{A, B\}$.

2. If $A$ and $B$ are adjacent in $\mathcal{P}$, there is an "−" (arrowtail) at the $A$ end of the edge iff $A$ is an ancestor of $B$ in every member of $\mathcal{M}$.

3. If $A$ and $B$ are adjacent in $\mathcal{P}$, there is an ">" (arrowhead) at the $B$ end of the edge iff $B$ is not an ancestor of $A$ in every member of $\mathcal{M}$.

4. If $A$ and $B$ are adjacent in $\mathcal{P}$, an "o" at the $A$ end of the edge entails that in some DAG in $\mathcal{M}$, $A$ is an ancestor of $B$ and in some other DAG in $\mathcal{M}$, $A$ is not an ancestor of $B$. (In (Richardson & Spirtes, 2002), this is called a maximally oriented PAG.)

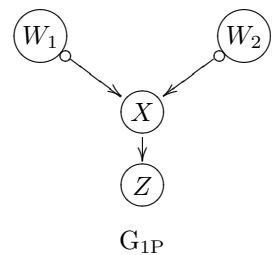

$G_{1P}$

Figure 6: A Y PAG.

For example, suppose $\mathcal{M}$ is the Markov equivalence class of the Y structure. It can be shown that the PAG that represents a Y structure is in Figure 6, indicating that for every DAG in $\mathcal{M}$, the following conditions hold:

- $X$ is not an ancestor of $W_1$ or $W_2$.
- $Z$ is not an ancestor of $X$.
- $X$ is an ancestor of $Z$.
- $W_1$ and $W_2$ are ancestors of $X$ in some members of $\mathcal{M}$, and not ancestors of $X$ in other members of $\mathcal{M}$.

---

[6]The results in (Nishii, 1988) are based on "almost surely" convergent proofs, which guarantee that in the large sample limit the data will with probability 1 support the stated convergence.

[7]If there are several Bayesian networks that contain the fewest number of parameters that can represent the data generating distribution, then the result states that the sum of their posterior probabilities is equal to 1.

**Definition 5 (DAG PAG).** For a PAG $\mathcal{P}$, if there is an assignment of arrowheads and arrowtails to the "o" endpoints in $\mathcal{P}$ such that the resulting graph is a DAG that has the same d-separation relations as $\mathcal{P}$, then $\mathcal{P}$ is a DAG PAG.

For example, a Y PAG is a DAG PAG because the DAG in Figure 5 (which we will call the Y DAG) has the same d-separation relations as the PAG in Figure 6. A DAG PAG can be parameterized in the same way as a corresponding DAG. Every DAG PAG has the same d-separations over the measured variables as the DAGs that it represents. Every DAG PAG can be assigned a score equal to the score of any of the DAGs that it represents. The reader is referred to (Spirtes et al., 1999) and (Spirtes et al., 2000, pages 299–301) for additional details about PAGs. The FCI algorithm is a constraint-based algorithm that generates a PAG from data faithful to a DAG represented by the PAG. A listing of the PAGs over four variables that includes the Y PAG with a discussion of their causal implications is presented in (Richardson & Spirtes, 2003).

**Definition 6 (Embedded pure Y structure).** Let B be a causal Bayesian network with structure S. We say that B contains an *embedded pure* Y structure (EPYS) involving the variables $W_1, W_2, X$ and $Z$, iff all and only the following d-separation conditions hold among the variables $W_1, W_2, X$ and $Z$ ($A > < B|\mathbf{C}$ means that $A$ and $B$ are d-separated conditioned on $\mathbf{C}$):

- $W_1 > < W_2|\{\}$; $W_1 > < Z|\{X\}$
- $W_1 > < Z|\{X, W_2\}$; $W_2 > < Z|\{X\}$
- $W_2 > < Z|\{X, W_1\}$

**Context 1.** Let B be a complete-table Bayesian network involving the variables $W_1, W_2, X$ and $Z$. Furthermore, let B be the data generating model for data on just $W_1, W_2, X$ and $Z$. In general, B may contain other variables, which we consider as hidden with regard to the data being measured on these four variables.

Let $\theta_{w1w2xz}$ be the data generating distribution on the variables $W_1, W_2, X$ and $Z$ that is given by a marginal distribution of B. Suppose that every d-separation condition among $W_1, W_2, X$ and $Z$ in B implies a corresponding independence according to $\theta_{w1w2xz}$ (call this the *marginal Markov condition*)[8]. Suppose also that every d-connection condition among $W_1, W_2, X$ and $Z$ implies a corresponding dependence according

---
[8]The marginal Markov condition is entailed by the Markov condition, and is strictly weaker than the Markov condition.

to $\theta_{w1w2xz}$ (call this the *marginal Faithfulness condition*)[9].

To summarize:

1. Let $S_y$ denote the Y structure in Figure 5.
2. Let $\mathbf{V}_{Sy}$ denote the variables in $S_y$.
3. Let B be the Bayesian network generating the data.
4. Let $\mathbf{V}_B$ denote the variables in B.
5. In general $\mathbf{V}_{Sy} \subseteq \mathbf{V}_B$.
6. Assume the marginal Markov and Faithfulness conditions hold for B with respect to $\theta_{w1w2xz}$.

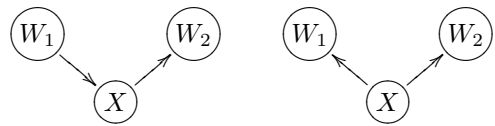

Figure 7: $X$ is a non-collider in both the models

**Definition 7 (Non-collider).** A variable $X$ is said to be a non-collider on a path if it does not have two incoming arcs (arrowheads). See Figure 7 for examples.

**Lemma 3.** *If* B *contains an EPYS, then in the large sample limit, the Y PAG receives a higher score than any other DAG PAG over the same variables, with probability 1.*

*Proof.* If B contains an EPYS, then by the Marginal Faithfulness condition, the marginal population distribution is faithful to the Y PAG, and hence to the Y DAG. By Theorem 3, the Y PAG receives a higher score than any other DAG over the same variables, and hence any other DAG PAG. □

Let PAG(B) be the PAG for Bayesian network(B) over the set of measured variables O.

**Lemma 4.** *If* B *does not contain an EPYS, and the population distribution is faithful to PAG*(B)*, then in the large sample limit, with probability 1 there is a DAG PAG* $\mathcal{P}$ *over* O *that receives a higher score than the Y PAG.*

*Proof.* Suppose that B does not contain an EPYS. There are three cases.

Suppose first that PAG(B) contains any adjacency between some pair of observed vertices $A$ and $C$ that are not adjacent in the Y PAG. It follows that $A$ and $C$ are d-separated conditional on some subset of **O** in the Y PAG but not in PAG(B). Hence, by the Marginal Faithfulness condition, in the population distribution $A$ and $C$ are dependent conditional on every subset of **O**. Hence the Y PAG cannot represent the marginal population distribution. Some DAG can represent the marginal population distribution, since a DAG in which every pair of vertices are adjacent can represent any distribution. Hence there is a DAG G with the fewest number of parameters that can represent the marginal population distribution. By Lemma 2, in the large sample limit with probability 1, the Y PAG receives a lower score than G.

---
[9]The marginal Faithfulness condition is entailed by the Faithfulness condition, and is strictly weaker than the Faithfulness condition.

Suppose next that PAG(B) contains the same adjacencies as the Y PAG, but different orientations. It is easy to see by exhaustively considering all of the possible PAGs with the same adjacencies as the Y PAG that each of them is a DAG PAG. Since the two different PAGs have the same d-separation relations as some pair of different DAGs over **O**, we can reason about their d-separation relations using two DAGs. Given that they have the same adjacencies, the Y DAG and the DAG G with the same d-separation relations as PAG(B) are not equivalent iff they have different unshielded colliders. Suppose that there is some unshielded collider $A \rightarrow D \leftarrow C$ in the Y DAG, but not in G. It follows that $A$ and $C$ are d-separated conditional on some subset of variables in the Y DAG, and that every set that d-separates $A$ and $C$ in the Y DAG does not contain D. In G, $A$ and $C$ are d-separated conditional only on subsets of variables that do contain D. Hence there is a d-separation relation in the Y PAG that is not in G. The case where $A \rightarrow D \leftarrow C$ is an unshielded collider in G, but not in the Y DAG, is analogous. Hence, by the marginal Faithfulness assumption the Y PAG cannot represent the marginal population distribution. By Lemma 2, in the large sample limit with probability 1, the Y PAG receives a lower score than some other DAG with the fewest parameters that does represent the population distribution.

Finally, suppose that PAG(B) contains a proper subset of the adjacencies in the Y PAG. Inspection shows that all of the PAGs with subsets of adjacencies of the Y PAG are DAG PAGs. Hence, by the Marginal Faithfulness condition, the population distribution is faithful to some such PAG, and hence faithful to a corresponding DAG. By Theorem 3 the DAG that the population distribution is faithful to receives a higher score than the Y PAG with probability 1 in the large sample limit. □

Let us consider the following example. Assume that B contains the substructure shown in Figure 8. We refer to this structure as Near-Y DAG or N-Y DAG

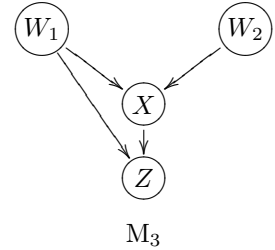

$M_3$

Figure 8: A Near-Y structure.

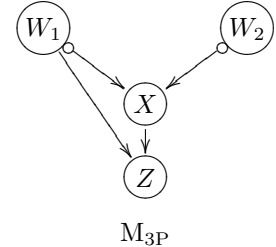

$M_{3P}$

Figure 9: A Near-Y PAG.

for short. The N-Y DAG has an additional arc from $W_1$ to $Z$ when compared to the Y DAG shown in Figure 5.[10] Note that for the N-Y DAG the d-separation $W_1 \mathbin{><} Z | \{X\}$ does not hold and hence if B contains an N-Y DAG, it will not be an embedded pure Y structure. According to Lemma 4 then with probability 1 there is a DAG PAG $\mathcal{P}$ over **O** that receives a higher score than the Y PAG. Such a DAG PAG $\mathcal{P}$ is shown in Figure 9.

The N-Y DAG also has interesting independence/dependence properties. It is the only member of its independence class over the observed variables $W_1$, $W_2$, $X$, and $Z$. A causal claim can be made for the arc from $X$ to $Z$ in the N-Y DAG and it can be estimated from $P(x | z, w)$. We plan to generate a formal proof of causality for the N-Y DAG as part of our future work.

**Theorem 4.** *Assume that Context 1 holds. In the large sample limit, in scoring DAGs on $\mathbf{V}_{Sy}$, $S_y$ is assigned the highest score, iff* B *contains a corresponding EPYS, and if* B *contains such an EPYS, then $X$ is an ancestor of $Z$ in* B, *and there is no unmeasured common cause of $X$ and $Z$.*

*Proof.* By Lemmas 3 and 4, $S_y$ is assigned the highest score, iff B contains a corresponding EPYS. In (Spirtes et al., 2000) it is shown that if there is a directed edge from $X$ to $Z$ in a PAG (as in the Y PAG), then $X$ is an ancestor of $Z$ in every DAG represented by the

---

[10]The N-Y DAG can have either the extra arc from $W_1$ to $Z$ or the arc from $W_2$ to $Z$, but not both.

PAG, and there is no hidden common cause of $X$ and $Z$ in any such DAG. □

Theorem 4 indicates that local Bayesian causal discovery using Y structures is possible (under assumptions), even when the data generating process is assumed to be a causal Bayesian network with hidden variables.

## 5 Discussion

The local Bayesian causal discovery based on Y structures may have practical applications on large data sets such as gene expression data sets with thousands of variables or large population-based data sets with hundreds of thousands of records. As the Y structure represents an unconfounded causal influence of a variable $X$ on variable $Z$, it can be used for performing planned interventions leading to desirable outcomes. When experimental studies are contra-indicated, due to ethical, logistical, or cost considerations, causal discovery from observational data remains the only feasible approach. Moreover, in resource limited settings such methods can be initially used to generate plausible causal hypothesis that can then be tested using experimental methods resulting in better utilization of available resources. As part of our future work, we plan to relax some of our causal discovery assumptions (for example, the acyclicity assumption) and extend the proofs to more complex Y structures such as those with measured confounders.

## A Appendix: Search

The proofs presented in this paper for the EPYS do not depend on a particular search heuristic for the identification of tetrasets (sets of four variables) for Y structure scoring. An obvious search method is to search for EPYS in all possible tetrasets of a domain. For different finite sample sizes the BLCD search method has been shown to be effective in simulation studies. The BLCD search first estimates the Markov blanket of a node $X$, and creates sets of four variables by choosing three nodes from the MB of $X$ in addition to $X$. The tetrasets are scored using the Score function described in Appendix B. A preliminary version of the BLCD algorithm was published in (Mani & Cooper, 2004). The reader is referred to (Mani, 2005) for additional details, including results of an extensive set of simulation experiments.

## B Appendix: Scoring measure

The Score function assigns a score to a model that represents the probability of the model given data and prior knowledge. For scoring the DAGs, we use the Bayesian likelihood equivalent (BDe) scoring measure (Heckerman et al., 1995). We use uniform, non-informative priors in order to test the ability of the algorithm to discover causal relationships from data, rather than from a combination of data and prior knowledge. The latter introduces two sources of experimental variation. Note that for a Y structure, the causal claim is valid for only the arc from $X$ to $Z$. We represent the Y structure using the notation $X \Rightarrow Z$. In the large sample limit under the causal Markov and causal Faithfulness assumptions (see Section 2), $P(X \Rightarrow Z|D)$ (D denotes the dataset) can be estimated using Equation 3:

$$\frac{\text{Score}(G_1|D)}{\sum_{i=1}^{543} \text{Score}(G_i|D)} \quad (3)$$

where $G_i$ represents one of the 543 CBNs over $\mathbf{V} = \{W_1, W_2, X, Z\}$.

Note that $P(X \Rightarrow Z|D)$ is a heuristic approximation to what would be obtained if we were to explicitly perform a full Bayesian scoring that includes hidden variables. The score for each of the 543 measured structures represents a score for both the measured structure itself, as well as the score for an infinite number of structures with hidden variables. Interpreting the score as a probability is a heuristic approximation borne out by simulation studies (Mani, 2005), but the theorems in the paper do not depend upon that approximation. The only probabilities being considered are the probability of being a Y structure and the probability of not being a Y structure. The theorems in the paper show the conditions under which $P(X \Rightarrow Z|D)$ converges to the correct value in the large sample limit. In the large sample limit the posterior probability of $G_1$ will be greater relative to the other 542 models if indeed (1) $X$ causally influences $Z$ in an unconfounded manner, (2) $X$ is an unshielded collider of $W_1$ and $W_2$ in the distribution of the causal process generating the data, and (3) the Markov and Faithfulness conditions hold.

**Acknowledgements**

We thank the anonymous reviewers for helpful suggestions. We also thank Thomas Richardson for comments on an earlier version of this paper. This research was supported in part by grant IIS-0325581 from the National Science Foundation and by grant R01-LM008374 from the National Library of Medicine awarded to Greg Cooper.